# PythonRobotics: a Python code collection of robotics algorithms

Atsushi Sakai　　　　　　　　　　　　　　　　　University of California, Berkeley
https://atsushisakai.github.io/

Daniel Ingram　　　　　　　　　　　　　　　　　Appalachian State University
ingramds@appstate.edu

Joseph Dinius　　　　　　　　　　　　　　　　　inVia Robotics, Inc.
jdinius@inviarobotics.com

Karan Chawla　　　　　　　　　　　　　　　　　Skyryse Inc.
karan@skyryse.com

Antonin Raffin　　　　　　　　　　　　　　　　　ENSTA ParisTech
antonin.raffin@ensta-paristech.fr

Alexis Paques　　　　　　　　　　　　　　　　　Unmanned Life
Alexis@unmanned.life

**Abstract**

This paper describes an Open Source Software (OSS) project: PythonRobotics[21]. This is a collection of robotics algorithms implemented in the Python programming language. The focus of the project is on autonomous navigation, and the goal is for beginners in robotics to understand the basic ideas behind each algorithm. In this project, the algorithms which are practical and widely used in both academia and industry are selected. Each sample code is written in Python3 and only depends on some standard modules for readability and ease of use. It includes intuitive animations to understand the behavior of the simulation.

# 1　Introduction

In recent years, autonomous navigation technologies have received huge attention in many fields. Such fields include, autonomous driving[22], drone flight navigation, and other transportation systems.

An autonomous navigation system is a system that can move to a goal over long periods of time without any external control by an operator. The system requires a wide range of technologies: It needs to know where it is (localization), where it is safe (mapping), where and how to move (path planning), and how to control its motion (path following). The autonomous system would not work correctly if any of these technologies is missing.

Educational materials are becoming more and more important for future developers to learn basic autonomous navigation technologies. Because these autonomous technologies need different technological skill sets such as: linear algebra, statistics, probability theory, optimization theory, and control theory etc. It needs a lot of time to be familiar with these technological areas. Therefore, good educational resources for learning basic autonomous navigation technologies are needed. Our project which is described in this paper aims to be one such resource.

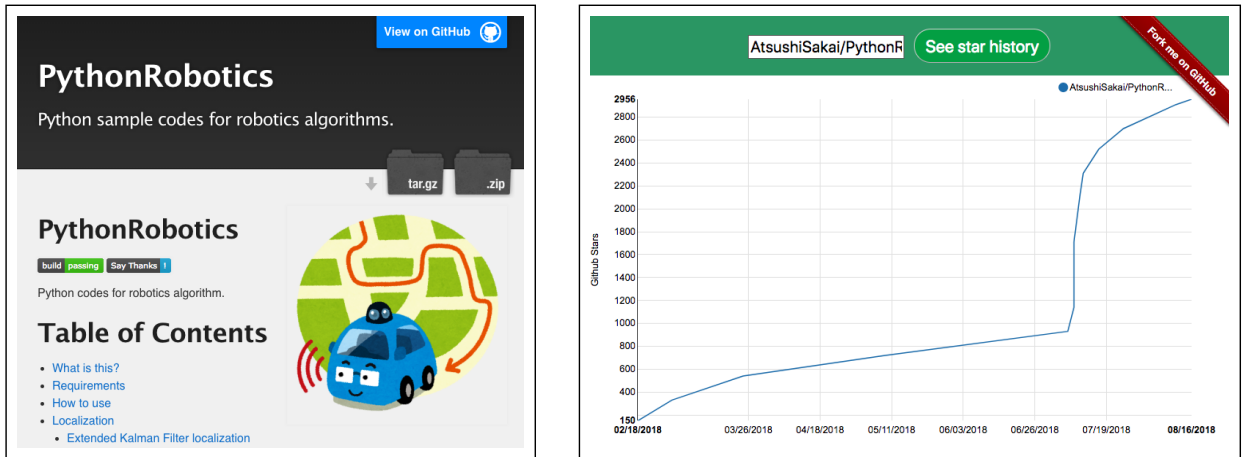

Figure 1: Left: PythonRobotics Project page, Right: GitHub star history graph using [16]

In this paper, an Open Source Software(OSS) project: PythonRobotics[21] is described. This project provides a code collection of robotics algorithms, especially focusing on autonomous navigation. The principle goal is to provide beginners with the tools necessary to understand it. It is written in Python[12] under MIT license[7]. It has a lot of simulation animations that shows behaviors of each algorithm. It helps for learners to understand its fundamental ideas. The readers can find the animations in the project page https://atsushisakai.github.io/PythonRobotics/. The left figure in Fig.1 shows the front image of the project page. All source codes of this project are provided at https://github.com/AtsushiSakai/PythonRobotics.

This project was started from Mar. 21 2016 as a self-learning project. It already has over 3000 stars on GitHub, and the right figure in Fig.1 shows the history graph of the star counts using [16].

This paper is organized as follows: Section 2 reviews the related works. The philosophy of this project is presented in Section 3. The repository structure and some technical backgrounds and simulation results are provided in Section 4. Conclusions are drawn and some future works are presented in Section 5. Section 6 acknowledges for contributors and supporters.

## 2 Related works

There are already some great educational materials for learning autonomous navigation technologies.

S. Thrun et al. wrote a great textbook "Probabilistic robotics" which is a bible of localization, mapping, and Simultaneous Localization And Mapping (SLAM) for mobile robotics[31]. E. Frazzoli et al. wrote a great survey paper about path planning and control techniques for autonomous driving [25]. G. Bautista et al. wrote a survey paper focusing on path planning for automated vehicles[22]. J. Levinson wrote an overview paper about systems and algorithms towards fully autonomous driving[24].

These papers help readers to learn state-of-the-art autonomous navigation technologies. However, it might be difficult for beginners to understand the basic ideas of the technologies and algorithms because the papers don't include implementation examples.

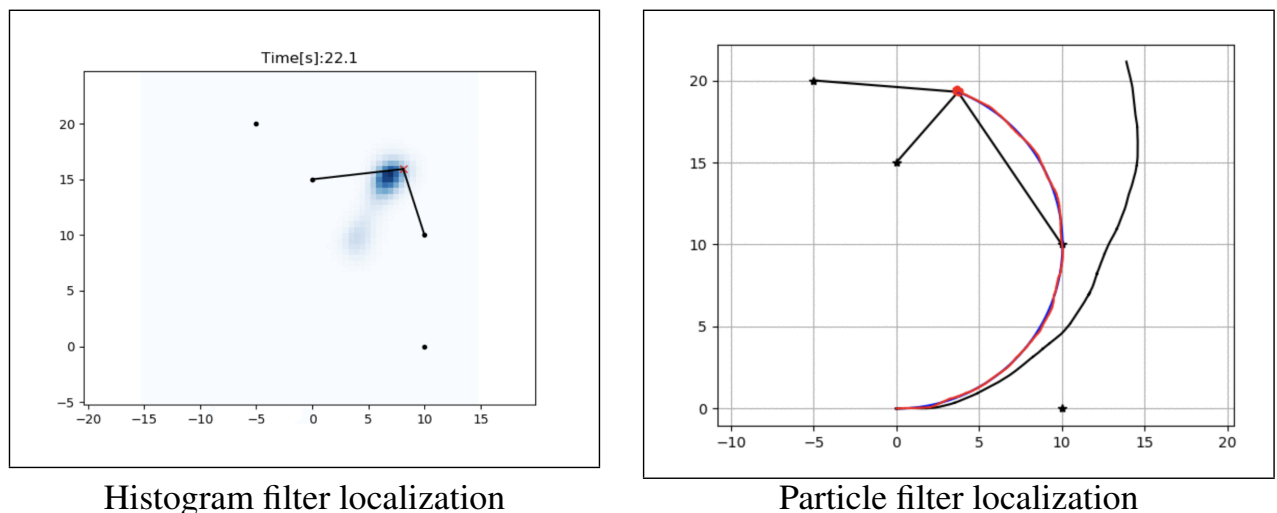

Histogram filter localization    Particle filter localization

Figure 2: Localization simulation results

Many universities provide great classes to learn robotics and share their lecture notes. For example, University of Freiburg provides an introduction class to mobile robotics[2]. Swiss Federal Institute of Technology in Zurich (ETH in Zurich) also provides a class about autonomous mobile robot[3] and robotics programming with Robot Operating System(ROS)[4].

Like the academic literature referenced, these lecture notes also help readers to learn autonomous navigation technologies. However, it might be also difficult for beginners of robotics to understand the basic ideas of the technologies and algorithms if the reader is not a student of the class, because these lecture notes doesn't include actual implementation examples of robotics algorithms.

Robot Operating System (ROS) is a middle-ware for robotics software development[13][28]. ROS initially released in 2007, and it has become the defacto standard platform in robotics software development. ROS includes basic navigation software such as the Adaptive Monte Carlo Localization (AMCL) package and the local path planner based on Dynamic Window Approach, etc[14]. Many autonomous navigation packages using ROS are also open-sourced.

ROS packages can also be useful to learn autonomous navigation algorithms, however the documentation might be inconsistent and it can be difficult to find implementation details about the algorithms used. ROS packages are usually written in C++ because the focus is computational efficiency and not readability, which makes learning from the code harder.

Udacity Inc, which is an online learning company founded by S. Thrun, et al, is providing great online courses for autonomous navigation and autonomous driving [17]. These courses provide not only technical lectures, but also programming homework using Python. This leads to a great learning strategy for beginners to understand the technical background required for autonomous systems.

The PythonRobotics project seeks to extend Udacity's approach by giving beginners the necessary tools to understand and make further study into the methods of autonomous navigation. The details of concepts of this OSS project will be described in the next section.

## 3   Philosophy

In this section, the philosophy of this project is described. The PythonRobotics project is based on three main philosophies. Each philosophy will be discussed separately in this section.

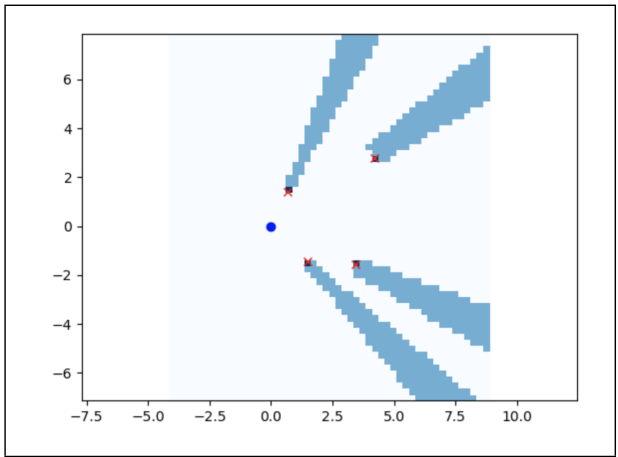 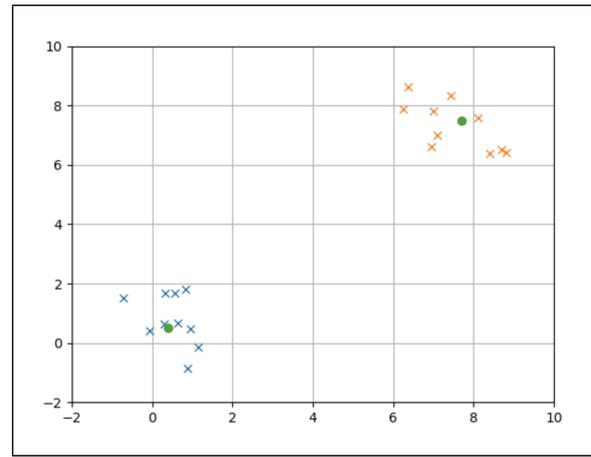

| Grid mapping with 2D ray casting | 2D object clustering with k-means algorithm |

Figure 3: Mapping simulation results

## 3.1 Readability

The first philosophy is that the code has to be easy to read. If the code is not easy to read, it would be difficult to achieve our goal of allowing beginners to understand the algorithms. Python[12] programming language is adopted in this project. Python has great libraries for matrix operation, mathematical and scientific operation, and visualization, which makes code more readable because such operations don't need to be re-implemented. Having the core Python packages allows the user to focus on the algorithms, rather than the implementations.

## 3.2 Practicality

The second philosophy is that implemented algorithms have to be practical and widely used in both academia and industry. We believe learning these algorithms will be useful in many applications. For example, Kalman filters and particle filter for localization, grid mapping for mapping, dynamic programming based approaches and sampling based approaches for path planning, and optimal control based approach for path tracking. These algorithms are implemented in this project.

## 3.3 Minimal dependencies

The last philosophy is minimal dependencies. Having few external dependencies allows us to run code samples easily and to convert the Python codes to other programming languages, such as C++ or Java for more practical application. Each sample code only depends some modules on Python3 as bellow.

- numpy[8] for matrix and vector operations

- scipy[15] for mathematical, scientific computing

- matplotlib[6] for plotting and visualization

- pandas[10] for data import and manipulation.

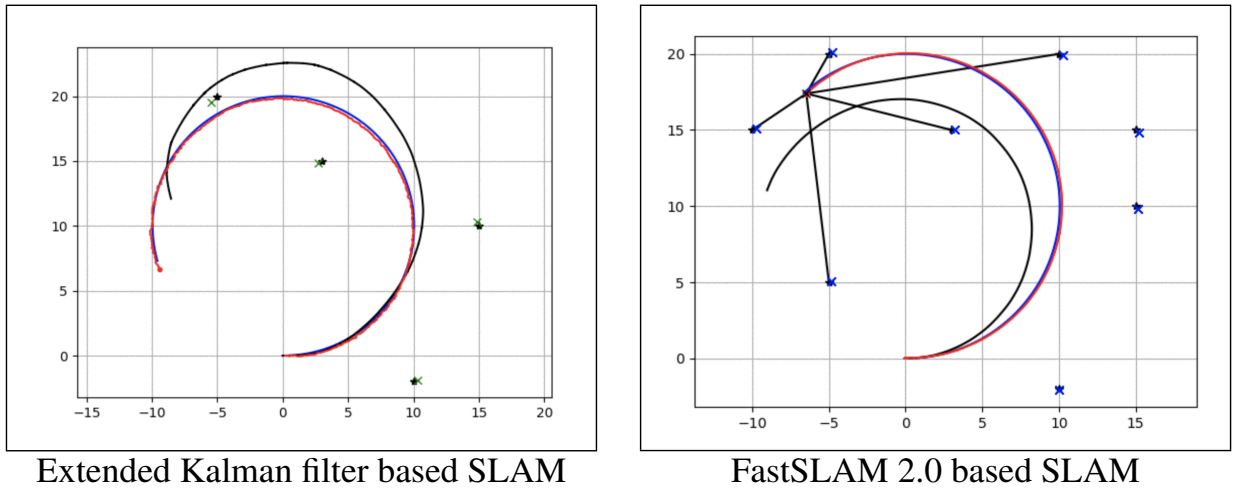

| Extended Kalman filter based SLAM | FastSLAM 2.0 based SLAM |

Figure 4: SLAM simulation results

- cvxpy[5] for convex optimization

These modules are OSS and can also be used for free. This repository software doesn't depend on any commercial software.

# 4 Repository structure

In this section, the brief repository structure is described.

There are five directories, each one corresponding to a different technical category in autonomous navigation: localization, mapping, SLAM, path planning, and path tracking. There is one sub-directory per algorithm, which has a sample code implementing and testing the algorithm. In the following subsections, some algorithms and some simulation examples in each category are described.

## 4.1 Localization

Localization is the ability of a robot to know its position and orientation with sensors such as Global Navigation Satellite System:GNSS etc. In localization, Bayesian filters such as Kalman filters, histogram filter, and particle filter are widely used[31]. Fig.2 shows localization simulations using histogram filter and particle filter.

## 4.2 Mapping

Mapping is the ability of a robot to understand its surroundings with external sensors such as LIDAR and camera. Robots have to recognize the position and shape of obstacles to avoid them. In mapping, grid mapping and machine learning algorithms are widely used[31][18]. Fig.3 shows mapping simulation results using grid mapping with 2D ray casting and 2D object clustering with k-means algorithm.

## 4.3 SLAM

Simultaneous Localization and Mapping (SLAM) is an ability to estimate the pose of a robot and the map of the environment at the same time. The SLAM problem is hard to solve, because a map is needed for localization and localization is needed for mapping. In this way, SLAM is often said to be similar to a 'chicken-and-egg' problem. Popular SLAM solution methods include the extended Kalman filter, particle filter, and Fast SLAM algorithm[31]. Fig.4 shows SLAM simulation results using extended Kalman filter and results using FastSLAM2.0[31].

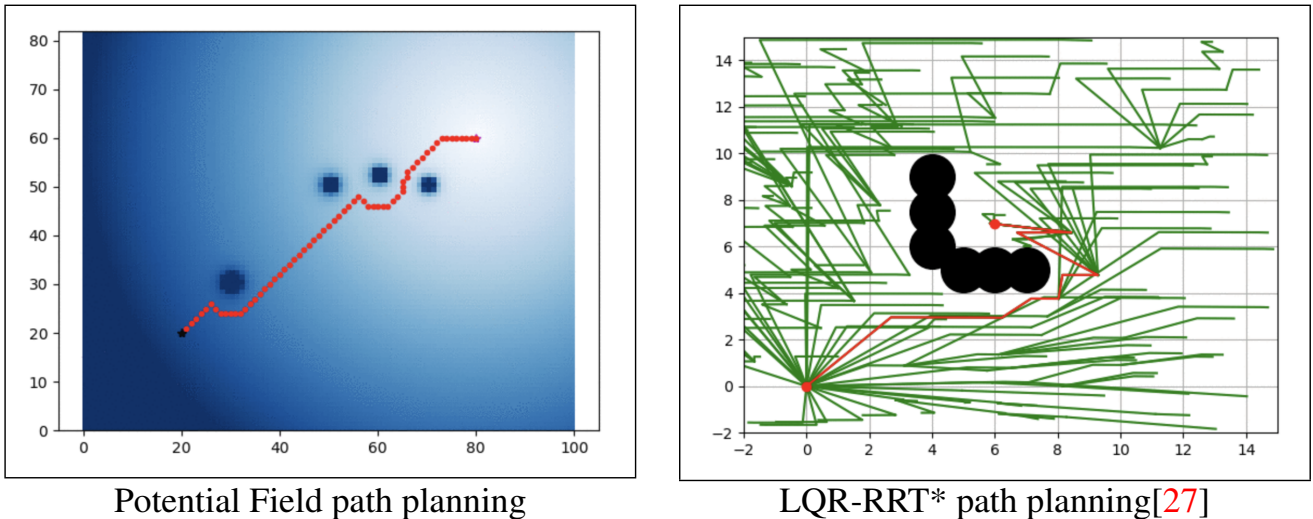

| Potential Field path planning | LQR-RRT* path planning[27] |

Figure 5: Path planning simulation results

## 4.4 Path planning

Path planning is the ability of a robot to search feasible and efficient path to the goal. The path has to satisfy some constraints based on the robot's motion model and obstacle positions, and optimize some objective functions such as time to goal and distance to obstacle. In path planning, dynamic programming based approaches and sampling based approaches are widely used[22]. Fig.5 shows simulation results of potential field path planning and LQR-RRT* path planning[27].

## 4.5 Path tracking

Path tracking is the ability of a robot to follow the reference path generated by a path planner while simultaneously stabilizing the robot. The path tracking controller may need to account for modeling error and other forms of uncertainty. In path tracking, feedback control techniques and optimization based control techniques are widely used[22]. Fig.6 shows simulations using rear wheel feedback steering control and PID speed control, and iterative linear model predictive path tracking control[27].

## 5 Conclusion and future work

In this paper, we introduced PythonRobotics, a code collection of robotics algorithms, with a focus on autonomous navigation. Related works of this project, some key ideas about this OSS project, and brief structure of this repository and simulation results were described. The future works for this project are as follows:

- Adding technical and mathematical documentation using Jupyter notebook[1].

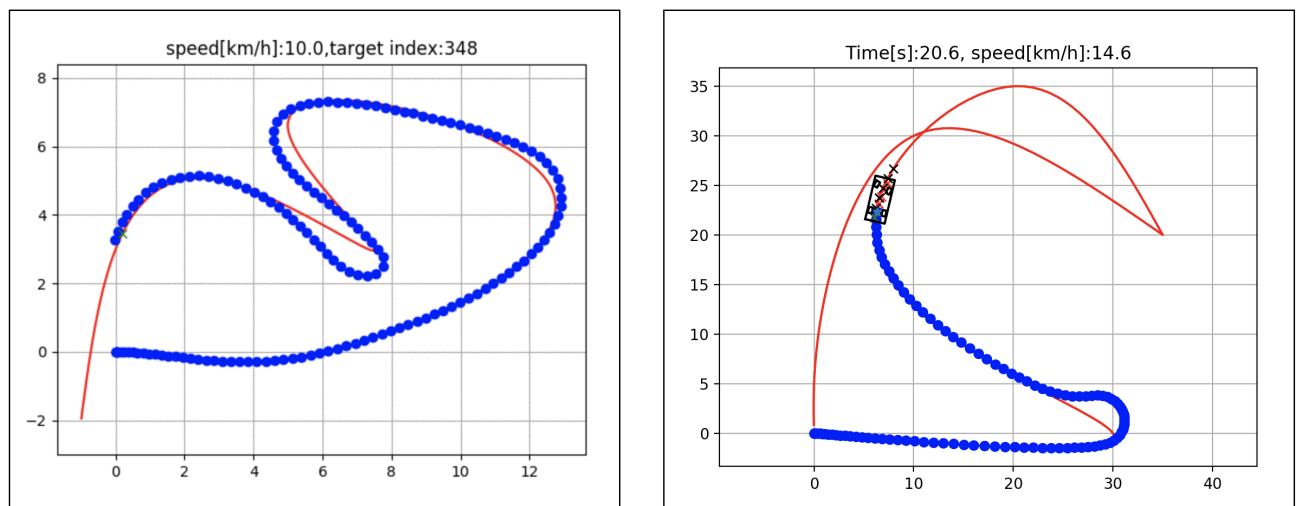

Rear wheel feedback steering control and PID speed control [25]

Iterative linear model predictive control

Figure 6: Path tracking simulation results

- Adding image processing samples for autonomous navigation using OpenCV[9].

- Adding simple multi-robot simulations for autonomous navigation.

- Adding a comparison to find which algorithm fits best for a specific application.

If readers are interested in these future projects, contributions are welcome. Readers can also support this project financially via Patreon[11].

# 6 Acknowledgments

We appreciate all contributors: Atsushi Sakai[30], Daniel Ingram[23], Joe Dinius[20], Karan Chala[19], Antonin RAFFIN[29], and Alexis Paques[26]. We also appreciate all robotics lovers who give a star to this repository in GitHub.

# References

[1] Project jupyter. http://jupyter.org/.

[2] Introduction to mobile robotics - ss 2018 - arbeitsgruppe: Autonome intelligente systeme, . http://ais.informatik.uni-freiburg.de/teaching/ss18/robotics/.

[3] Autonomous mobile robots - spring 2018 autonomous systems lab | eth zurich, . http://www.asl.ethz.ch/education/lectures/autonomous_mobile_robots/spring-2018.html.

[4] Programming for robotics - ros âĂŞ robotic systems lab eth zurich, . http://www.rsl.ethz.ch/education-students/lectures/ros.html.

[5] Cvxpy. http://www.cvxpy.org/index.html.

[6] Matplotlib. https://matplotlib.org/.

[7] Mit license. https://opensource.org/licenses/MIT.

[8] Numpy. http://www.numpy.org/.

[9] Opencv. https://opencv.org/.

[10] pandas. https://pandas.pydata.org/.

[11] Atsushi sakai is creating open source software | patreon. https://www.patreon.com/myenigma.

[12] Python. https://www.python.org.

[13] Ros (robot operating system), . http://wiki.ros.org.

[14] Ros navigation stack, . http://wiki.ros.org/navigation.

[15] Scipy. https://www.scipy.org/.

[16] Star history. http://www.timqian.com/star-history/#AtsushiSakai/PythonRobotics.

[17] Artificial intelligence for robotics | udacity. https://www.udacity.com/course/artificial-intelligence-for-robotics--cs373.

[18] Christopher M. Bishop. *Pattern Recognition and Machine Learning (Information Science and Statistics)*. Springer-Verlag, Berlin, Heidelberg, 2006. ISBN 0387310738.

[19] Karan Chawla. Github account. https://github.com/karanchawla.

[20] Joe Dinius. Github account. https://github.com/jwdinius.

[21] Atsushi Sakai et al. Atsushisakai/pythonrobotics: Python sample codes for robotics algorithms. https://github.com/AtsushiSakai/PythonRobotics.

[22] David Gonzalez Bautista, JoshuÃl' PÃl'rez, Vicente Milanes, and Fawzi Nashashibi. A review of motion planning techniques for automated vehicles. pages 1–11, 11 2015.

[23] Daniel Ingram. Github account. https://github.com/daniel-s-ingram.

[24] Jesse Levinson, Jake Askeland, Jan Becker, Jennifer Dolson, David Held, Soeren Kammel, J. Zico Kolter, Dirk Langer, Oliver Pink, Vaughan Pratt, Michael Sokolsky, Ganymed Stanek, David Stavens, Alex Teichman, Moritz Werling, and Sebastian Thrun. Towards fully autonomous driving: systems and algorithms. In *Intelligent Vehicles Symposium (IV), 2011 IEEE*, 2011.

[25] Brian Paden, Michal Cap, Sze Zheng Yong, Dmitry Yershov, and Emilio Frazzoli. A survey of motion planning and control techniques for self-driving urban vehicles. 2016.

[26] Alexis Paques. Github account. https://github.com/AlexisTM.

[27] Alejandro Perez, Robert Platt Jr, George Konidaris, Leslie P. Kaelbling, and Tomas Lozano-Perez. Lqr-rrt*: Optimal sampling-based motion planning with automatically derived extension heuristics. pages 2537–2542, 05 2012.

[28] Morgan Quigley, Ken Conley, Brian P. Gerkey, Josh Faust, Tully Foote, Jeremy Leibs, Rob Wheeler, and Andrew Y. Ng. Ros: an open-source robot operating system. In *ICRA Workshop on Open Source Software*, 2009.

[29] Antonin RAFFIN. Github account. https://github.com/araffin.

[30] Atsushi Sakai. Github account. https://github.com/AtsushiSakai.

[31] Sebastian Thrun, Wolfram Burgard, and Dieter Fox. *Probabilistic Robotics (Intelligent Robotics and Autonomous Agents)*. The MIT Press, 2005. ISBN 0262201623.